# Development of modeling and control strategies for an approximated Gaussian process

Shisheng Cui[1] and Chia-Jung Chang[1]

**Affiliation:**

[1]Harold and Inge Marcus Department of Industrial and Manufacturing Engineering

The Pennsylvania State University

University Park, PA 16802, USA

E-mail: suc256@psu.edu; cuc28@engr.psu.edu

**Corresponding author**:

Shisheng Cui

355 Leonhard Building

The Pennsylvania State University

University Park, PA 16802, USA

Tel: +1 (650) 804-2926

E-mail: suc256@psu.edu

# Development of modeling and control strategies for an approximated Gaussian process

**Abstract**

The Gaussian process (GP) model, which has been extensively applied as priors of functions, has demonstrated excellent performance. The specification of a large number of parameters affects the computational efficiency and the feasibility of implementation of a control strategy. We propose a linear model to approximate GPs; this model expands the GP model by a series of basis functions. Several examples and simulation studies are presented to demonstrate the advantages of the proposed method. A control strategy is provided with the proposed linear model.

INTRODUCTION

The Gaussian process (GP) is a powerful modeling tool that has many applications in research and practice. It provides a practical and probabilistic approach to learning in kernel machines. The GP is extensively applied as a prior of a true function. It typically enables an excellent fit to training data and can compute with an infinite set of possible functions in finite time to make reasonable predictions.

This paper considers the basic GP model and its corresponding linear regression approximation. Assume a set of inputs and outputs and $n$ datapoints. The inputs are real-valued vectors denoted by $x$, whereas the outputs are real-valued scalars denoted by $y$. To learn the most accurate function from the training set, the true function is considered to be a GP model. The model that we employ is

$$y = f(x) + \epsilon \tag{1}$$

where the observation $y$ consists of the true function value $f(x)$ and random noise $\epsilon \sim Nor(0, \sigma^2)$. The true function is assumed to be a GP

$$f(x) \sim GP(0, \tau^2 \psi) \tag{2}$$

where $\psi(x, x')$ is the correlation function that is defined as $cor(f(x), f(x'))$. A common choice for the correlation function is given by

$$\psi(x, x') = \exp\left\{-\sum_{i=1}^{d} k_i (x_i - x_i')^2\right\} \tag{3}$$

where $d$ is the dimension of input $x$ and $k = (k_1, k_2, \ldots, k_d)$ represents the correlation parameters. Thus, a GP can be completely characterized by a covariance function. Given the unknown function $f(x)$, the corresponding covariance function should be selected to reflect detailed prior knowledge about $f(x)$. Kennedy and O'Hagan (2001) presented general rules for specifying a covariance function.

Given a mean and covariance function, a GP is determined. However, the fitting may lose efficiency for high-dimensional data because correlations between any pair of points must be estimated and these computations can be time consuming. Another shortcoming is the difficulty of implementing this approach in a control strategy. In a common control scenario, our task is to calculate an appropriate input at each step to achieve the target output, which requires that we sequentially solve the following predictive equation for the desired input $x^*$:

$$y^* = \tau^2 \psi(x^*, x)^T (\tau^2 \Psi + \sigma^2 I_n)^{-1} y \qquad (4)$$

where $y^*$ is the target output value, $x$ is the vector of the training input, $y$ is the vector of the training outputs, $\Psi$ is the covariance matrix formed by the training inputs and $I_n$ is an identity matrix. Equation (4) is a regular predictive equation from GP regression. Because this equation is difficult to solve, the application of this method for direct process control is challenging. To overcome these drawbacks, this paper considers approximating the GP using a linear regression model. This approach can significantly reduce the parameters to be estimated, which simplifies the implementation of control strategies. Previous studies have not addressed linearizing the GP; thus, this study comprises a novel approach.

**LITERATURE REVIEW**

For the supervised learning problem, we must make assumptions about the characteristics of the underlying function. A common approach involves the development of a prior distribution for the observations. In this approach, the computations required for inference and learning are simplified for the GP. Thus, the GP is suitable for use as the prior distribution of these functions.

The main limitation is that the memory requirements and computational demands increase as the square and cube, respectively, with the number of training cases $nm$. Two common approaches are employed to overcome this limitation. In some papers, sparse approximation algorithms were employed (Candela and Rasmussen, 2005). Only a subset of the latent variables are treated as exact variables; the remaining variables are given some approximate but computationally inexpensive treatment. Another method to accelerate GP regression (Candela, Rasmussen and Williams, 2007) is to accelerate the matrix vector multiplication (MVM). To save time, the method provides an approximate solution instead of the exact solution that can be used to terminate the iteration at an earlier point. Several other methods (Chalupka, Williams and Murray, 2013) are available to accelerate the computation process.

If we employ GP regression in a control scheme, for example, adjusting the input values to achieve the target output, we will need an approximation method with fewer parameters to evaluate that easily computes the best inputs given a reference output. Inspired by Chang and Joseph (2013), we consider approximating the GP with a linear model. A variable selection method is needed to reduce the number of parameters (Breiman, 1995). With the approximating linear model, we can propose a predictive control strategy by minimizing the optimization cost function

$$J = w_1(y^* - \hat{y})^2 + w_2 \Delta x^2 \qquad (5)$$

$y^*$ is the reference variable

$\hat{y}$ is the current predictive variable

$\Delta x$ is the relative changes in inputs

$\omega_1$ and $\omega_2$ are weighting coefficients

which is a polynomial to compute the optimal inputs. Using the method from Parrilo and Sturmfels (2003), we can efficiently solve this optimization to ensure a suitable control strategy.

**METHODOLOGY**

Assume that the true function is a continuous function in the function space. It can be represented as a linear combination of some continuous basis functions

$$f(x) = \sum_{k=0}^{N} \omega_k \phi_k(x) \qquad (6)$$

where $\{\phi_0(x), \phi_1(x), \dots, \phi_N(x)\}$ are basis functions and $\omega_0, \omega_1, \dots \omega_N$ are the parameters to be measured. As Chang and Joseph (2013) suggested, the ability to represent the main and interaction effects of input $x$ should be carefully considered when selecting basis functions. Basis functions that can be employed to expand the GP model in this study include the Fourier series, the Wavelet series and Legendre polynomials. After selecting a basis function, the degree of the basis function and the order of the interaction effect should be determined. The number of basis functions that are required to represent the model is dependent on the true function to be approximated. We should initially use a large number and then set an appropriate number based on the result. Then, the model becomes

$$y = \sum_{k=0}^{N} \omega_k \phi_k(x) + \epsilon \qquad (7)$$

As a result, $N + 1$ parameters must be estimated; the computations are expensive for a large number of basis functions. Although some of the basis functions may not significantly affect the results, they can increase the variance and reduce the efficiency. Some variable selection strategies can be used to identify the important variables and estimate the corresponding parameters.

We consider the nonnegative garrote method (Breiman, 1995) due to its accuracy compared with subset selection and its ease of implementation compared with ridge regression. The procedure involves obtaining initial estimates and performing a constrained least squares optimization to obtain the final estimates by shrinking. Let $\tilde{\omega}_0, \tilde{\omega}_1, \ldots, \tilde{\omega}_N$ denote the initial estimates of the parameters and $c = \{c_0, c_1, \ldots, c_N\}$ denote the coefficients in the nonnegative garrote method. Then, $c$ is estimated by minimizing

$$\sum_{i=1}^{n}\left\{y_i - \sum_{k=0}^{N} c_k \tilde{\omega}_k \phi_k(x_i)\right\}^2 \tag{8}$$

subject to the constraints

$$\sum_{k=0}^{N} c_k \leq M$$

$$c_k \geq 0 \quad k = 0, 1, \ldots, N$$

where $M$ is a nonnegative real value that is used to shrink the parameters. When $M = 0$, the value of $c_k$ is 0. With an increase in $M$, some of the $c_k$s will become positive and the corresponding variables will be selected. With $c$ and $\omega$, the final linear prediction model is

$$\hat{y} = \sum_{k=0}^{N} c_k \tilde{\omega}_k \phi_k(x) \tag{9}$$

An important part of this study is to obtain initial estimates for the parameters that can be employed in the nonnegative garrote method; the progress achieved in this area is described in the next section.

To approximate the GP prior distribution, we obtain a prior for $\omega$ such that the distribution of $\sum_{k=0}^{N} \omega_k \phi_k(x)$ can approximate the GP prior and then approximate the posterior based on this prior, as suggested by Joseph (2006).

Let $\phi(x) = (\phi_0(x), \phi_1(x), \ldots, \phi_N(x))^T$. The problem is to obtain a $\omega$ such that $\phi(x)^T \omega \approx f(x)$ for every $f(x)$ in $\mathcal{X}$, which is the space of $x$ that contains all points. We use the least squares method to obtain $\phi(x)$. Then, we can obtain $\omega$ by minimizing

$$\int_{\mathcal{X}} \{f(x) - \phi(x)^T \omega\}^2 \, dx$$

where $\mathcal{X}$ denotes the space of $x$. The result is

$$\omega = \left\{ \int_{\mathcal{X}} \phi(x) \phi(x)^T dx \right\}^{-1} \int_{\mathcal{X}} \phi(x) f(x) dx \tag{10}$$

Assume that the GP model $GP(0, \tau^2 \psi)$ has continuous correlation functions $\psi(x, x')$ and that each basis function $\phi_k$ of the linear model $y = \sum_{k=0}^{N} \omega_k \phi_k(x) + \epsilon$ is integrable. The expectation of $\omega$ is

$$E(\omega) = E\left( \left\{ \int_{\mathcal{X}} \phi(x) \phi(x)^T dx \right\}^{-1} \int_{\mathcal{X}} \phi(x) f(x) dx \right)$$

$$= \left\{ \int_{\mathcal{X}} \phi(x) \phi(x)^T dx \right\}^{-1} \int_{\mathcal{X}} E(f(x)) \phi(x) dx = 0$$

The variance of $\omega$ is

$$Var(\omega) = E(\omega \omega^T)$$

$$= E\left( \left\{ \int_{\mathcal{X}} \phi(x) \phi(x)^T dx \right\}^{-1} \int_{\mathcal{X}} f(x) \phi(x) dx \int_{\mathcal{X}} f(x) \phi(x)^T dx \left\{ \int_{\mathcal{X}} \phi(x) \phi(x)^T dx \right\}^{-1} \right)$$

$$= \left\{ \int_{\mathcal{X}} \phi(x) \phi(x)^T dx \right\}^{-1} \int_{\mathcal{X}} \int_{\mathcal{X}} E(f(x) f(x')) \phi(x) \phi(x)^T dx dx' \left\{ \int_{\mathcal{X}} \phi(x) \phi(x)^T dx \right\}^{-1}$$

$$= \tau^2 \left\{ \int_{\mathcal{X}} \phi(x)\phi(x)^T dx \right\}^{-1} \int_{\mathcal{X}} \int_{\mathcal{X}} \psi(x,x')\phi(x)\phi(x)^T dx dx' \left\{ \int_{\mathcal{X}} \phi(x)\phi(x)^T dx \right\}^{-1} \quad (11)$$

Because the stochastic integral of a normally distributed random variable with a bounded variance also follows a normal distribution, we obtain

$$\omega \sim Nor(\mathbf{0}, \tau^2 \Sigma)$$

where $\mathbf{0}$ is a vector of $(N+1)$ zeroes and

$$\Sigma = \left\{ \int_{\mathcal{X}} \phi(x)\phi(x)^T dx \right\}^{-1} \int_{\mathcal{X}} \int_{\mathcal{X}} \psi(x,x')\phi(x)\phi(x)^T dx dx' \left\{ \int_{\mathcal{X}} \phi(x)\phi(x)^T dx \right\}^{-1} \quad (12)$$

In this study, Legendre polynomials are the basis functions. With Legendre polynomials, we obtain

$$\omega \sim Nor\left(0, \tau^2 \int_{\mathcal{X}} \int_{\mathcal{X}} \psi(x,x')\phi(x)\phi(x)^T dx dx'\right)$$

Assume that $\{\phi_k(x)\}$ is a set of Legendre polynomials, then

$$\int_{\mathcal{X}} \phi_i(x)\phi_j(x) dx = \begin{cases} 1 & if\ i = j \\ 0 & otherwise \end{cases}$$

We employ a simulation approach (Chang and Joseph, 2013) to compute the prior distribution of $\omega$. First, the set of points $S = \{x_1, x_2 \ldots, x_k\}$ is selected from $\mathcal{X}$. These points should fill in the space and $k$, which denotes the size of $S$, should be sufficiently large compared with $n$, which denotes the size of the data.

Let $\Phi_s$ denote the basis matrix that consists of the basis functions of each point in $S$, and let $\Psi$ denote the correlation matrix, where $\Psi_{ij} = \psi(x_i, x_j)$, $\{x_i\}$ is the set of points in $S$. Then, we obtain

$$\omega = (\Phi_s^T \Phi_s)^{-1} \Phi_s^T f \quad (13)$$

where $f = (f(x_1), f(x_2), \ldots, f(x_k))^T$. Using the approach for obtaining the estimated prior distribution of $\omega$, the following results are obtained:

$$E(\omega) = 0 \tag{14}$$

$$Var(\omega) = \tau^2 (\Phi_s^T \Phi_s)^{-1} \Phi_s^T \Psi \Phi_s (\Phi_s^T \Phi_s)^{-1} \tag{15}$$

Here, $\omega$ also follows a normal distribution

$$\omega \sim Nor\left(0, \tau^2 (\Phi_s^T \Phi_s)^{-1} \Phi_s^T \Psi \Phi_s (\Phi_s^T \Phi_s)^{-1}\right)$$

which has the optimal least squares approximation to the GP prior if $S$ is assumed to be the data space instead of $\mathcal{X}$.

Using the prior for $\omega$ that approximates the GP prior of $f(x)$, we are able to obtain the posterior for $\omega$. Because $y = f(x) + \epsilon$, $f(x) \sim GP(0, \tau^2 \psi)$, $\epsilon \sim Nor(0, \sigma^2)$, the posterior mean of $f(x)$ is given by

$$\tilde{f}(x) = \tau^2 \psi(x)^T (\tau^2 \Psi + \sigma^2 I_n)^{-1} y \tag{16}$$

Thus, the posterior mean of $\omega$ can be obtained as

$$\tilde{\omega} = \tau^2 \Sigma \Phi_D^T (\tau^2 \Phi_D \Sigma \Phi_D^T + \sigma^2 I_n)^{-1} y \tag{17}$$

where $\Phi_D$ is the $n \times (N+1)$ matrix generated from $\phi(x)^T \omega$ based on the points in the data and $\Sigma = (\Phi_s^T \Phi_s)^{-1} \Phi_s^T \Psi \Phi_s (\Phi_s^T \Phi_s)^{-1}$.

To select the significant variables, a series of minimization problems that are subject to nonnegative garrote constraints can be solved with a sequence of $M$ beginning with 0. Minimizing the root mean squared prediction error yields the best value of $M$

$$\text{RMSPE}(M) = \sqrt{\frac{1}{t}\sum_{i=1}^{t}\{y_i - \hat{y}_i\}^2} \tag{18}$$

where $(x_i, y_i)$ is the ith test point and $\hat{y}_i$ is the predicted value at $x_i$. Let $\hat{c}$ represent the estimates of the parameters at the best $M$. The linear model is

$$\hat{y} = \sum_{k=0}^{N} \hat{c}_k \tilde{\omega}_k \phi_k(x) \tag{19}$$

This linear model can be used to design a control strategy. Consider a predictive control scenario, in which we must calculate the optimal control moves to minimize the optimization cost function

$$J = \omega_1(y^* - \hat{y})^2 + \omega_2(x - x')^2 \tag{20}$$

where $y^*$ is the target value, $\hat{y}$ is our linear predictive model, $x$ is the input required for the calculations, $x'$ is the current input and $\omega_1$, $\omega_2$ are weight coefficients. The second part in the cost function is to ensure that the system is stable. It is equivalent to minimizing

$$J = \omega_1\left(y^* - \sum_{k=0}^{N} \hat{c}_k \tilde{\omega}_k \phi_k(x)\right)^2 + \omega_2(x - x')^2 \tag{21}$$

which is a polynomial optimization. This equation is actually a sum of squares (SOS) of polynomials. Let $p(x)$ be a polynomial of degree $2d$ and let $z$ be a vector of all monomials of the degree less than or equal to $d$. Then, $p(x)$ is SOS if and only if there exists $M$ such that

$$M \succcurlyeq 0$$

$$p(x) = z^T M z$$

where the number of components of $z$ is $\binom{n+d}{d}$. Because our objective $J(x)$ is a SOS, we have

$$K \succcurlyeq 0$$

$$J(x) = z^T K z$$

For each $q$ that satisfies the condition of $J(x) - q \geq 0$, we can construct a new matrix

$$K'_{ij} = \begin{cases} K_{ij} - q & if\ i = j = 1 \\ K_{ij} & otherwise \end{cases}$$

where $K_{ij}$s are elements of $K$. $J(x) - q = z^T K' z$, and $K' \succcurlyeq 0$. Thus, $J(x) - q$ is a SOS if and only if $J(x) - q \geq 0$. Our minimization problem is equivalent to finding the largest $q$ such that $J(x) - q$ is a SOS. Therefore, we can utilize the method of SOS optimization in the control strategy design.

## MODELING EXAMPLES

We approximate several functions using linear models and then present a control scheme with our method.

### Univariate Examples

The first four examples are univariate functions. The size of each training set is 100. We employ our approach to the training sets to learn linear approximating models. Eleven basis functions are employed in these processes. After obtaining the linear models, we compare them with the true functions to measure their performance by root mean squared prediction errors (RMSPEs) with a test set size of 10,000.

The four univariate functions that we employ are successively presented:

    a.   $f(x) = x^2$

Because this function is a polynomial, a linear model with polynomial basis functions should provide a reasonable approximation. Therefore, our approach should be suitable; we employ this function to start.

From the calculations, we obtain an approximated linear model

$$\hat{y} = 0.4607\phi_0(x) + 0.4278\phi_2(x)$$

with a RMSPE of 0.0035. As shown in Fig. 1, the true function and predictive function overlap, which indicates acceptable results.

[insert Fig. 1 here]

b. $f(x) = e^{4x}$

This function is selected because the function value changes rapidly when $x$ changes from 0 to 1. We determine whether our approach is capable of capturing this type of feature.

To approximate the true function, we assign a large value to $\tau$, which defines the covariance relation of the GP; $\tau$ is set to 1,000. Then, a linear model is obtained

$$\hat{y} = 9.6812\phi_0(x) + 12.5528\phi_1(x) + 9.4639\phi_2(x) + 5.2434\phi_3(x) + 2.2754\phi_4(x)$$
$$+ 0.8668\phi_5(x) + 0.2022\phi_6(x)$$

with a RMSPE of 0.0099. This linear model has more basis functions than the previous linear model because it is more complex. However, the order of this linear model is six, which ensures a simplified model and indicates that our approach is capable of approximating a complex univariate function with few polynomials. The predictive result is shown in Fig.2.

[insert Fig. 2 here]

c. $f(x) = x \sin \pi x$

This function has many optimums that are often difficult to learn. Thus, addressing this function using our approach is challenging, but we aim to achieve a reasonable approximation. As previously discussed, $\tau$ is set to 1,000, and we obtain the following linear model

$$\hat{y} = 0.4611\phi_0(x) + 0.1384\phi_2(x) - 0.2547\phi_4(x) + 0.0249\phi_6(x)$$

with a RMSPE of 0.0094. In the linear model, all basis functions are even functions, which capture the even features of the true function. As shown in Fig.3, the RMSPE indicates satisfactory results and that our approach is suitable for this type of function.

[insert Fig. 3 here]

To demonstrate the performance of our linear approximated model, we compare the predictive results of our approximated model to the predictive results of the original GP model. The predictive result is shown in Fig. 4. The RMSPE is 0.0187. Our linear model achieves a better result. This comparison verifies the strength of our approach.

[insert Fig. 4 here]

d. $f(x)$ is a GP model

In our approach, we assume that the prior distributions of the functions comprise a GP. Thus, we present an example, which is actually a GP model. First, a GP model is generated. The kernel function that is employed to characterize the model is the following squared exponential function:

$$K(x, x^{'}) = \exp\left(-\frac{|d|^2}{2l^2}\right)$$

Second, we obtain a linear approximated model with a RMSPE of 0.0112 which is shown in Fig.5. We achieve promising results because our approach assumes that the function has a GP prior distribution.

[insert Fig. 5 here]

With these four examples, we prove that our linear models can reasonable approximate the univariate functions, which can be useful in many applications. We consider other cases in the following paragraphs.

**Bivariate Example**

The univariate cases are acceptable but insufficient. We hope that our approach can also be applied to multivariate cases. We consider the true function as the bivariate Gaussian distribution function:

$$f(x) = \frac{1}{\sqrt{2\pi|\Sigma|}} \exp\left(-\frac{1}{2}(x-\mu)^T \Sigma^{-1}(x-\mu)\right)$$

Because the function contains two variates, we must define their interactions in the basis functions. The Legendre polynomials can be multiplied to define the interaction effect. The maximum order of the employed basis functions is six, which indicates a total of 28 basis functions. Thus, we obtain a slightly more complex approximated linear model

$$\hat{y} = 2.0940\phi_0(x_1)\phi_0(x_2) - 0.2763\phi_0(x_1)\phi_2(x_2) + 0.0383\phi_0(x_1)\phi_4(x_2)$$
$$+ 0.4808\phi_1(x_1)\phi_1(x_2) - 0.0928\phi_1(x_1)\phi_3(x_2) - 1.2000\phi_2(x_1)\phi_0(x_2)$$
$$+ 0.2642\phi_2(x_1)\phi_2(x_2) - 0.0129\phi_2(x_1)\phi_4(x_2) - 0.4282\phi_3(x_1)\phi_1(x_2)$$
$$+ 0.0691\phi_3(x_1)\phi_3(x_2) + 0.3377\phi_4(x_1)\phi_0(x_2) - 0.1156\phi_4(x_1)\phi_2(x_2)$$
$$+ 0.1335\phi_5(x_1)\phi_1(x_2) - 0.0924\phi_6(x_1)\phi_0(x_2)$$

with a RMSPE of 0.0565. The graph on the left in Fig. 6 depicts the distribution of the true function, whereas the graph on the right depicts the predictive function. The approximating function is similar to the true function. Thus, we can trust our approach from these promising results.

[insert Fig. 6 here]

This result demonstrates that this approach is also suitable for multivariate cases.

**Electric and Gas-Powered Vehicle Example**

Although the previous examples pertain to cases with continuous variables, we consider whether this approach is suitable for cases with discrete variables. We consider the problem of identifying the factors of the reasonable operation of an electric car or gasoline-powered vehicle in terms of variable fuel costs, given the uncertainty in retail power. The problem is addressed with a Monte Carlo simulation approach with DOE parameters and a stochastic component (empirical driving distance) for an eight-year simulation period. The four DOE parameters are mpkWh, mpg, gasoline cost and kWh cost. We are interested in the difference in the daily cost between gasoline options and electric powered options; the true function is

$$f(x) = x_1 \left( \frac{x_5}{x_3} - \frac{x_4}{x_2} \right)$$

$x_1$ = driving distance in miles (distance)

$x_2$ = miles per kilowatt-hour (mpkWh)

$x_3$ = miles per gallon (mpg)

$x_4$ = electricity price per kWh (utildol)

$x_5$ = gasoline price per gallon (gasdol)

where $x_2$, $x_3$, $x_4$ and $x_5$ represent the three-level discrete variates and $x_1$ is a continuous variate.

We consider the four discrete variates to be continuous variates but employ only three values in the training and testing sets. The training set contains 810 samples with ten samples from each of $3^4$ treatments. The testing set contains 236,520 points. We are only interested in the main effect and two-factor interactions. For the basis functions, we only consider the interception, the first order of the five variates and the second order of the interactions; the total number is 1+5+10=16.

Then, we obtain an approximated linear model:

$$\hat{y} = -0.0104 + 0.0477x_1 - 0.0778x_2 + 0.0260x_3 - 3.823x_4 + 0.4727x_5 - 0.00150x_6$$
$$+ 0.00531x_7 - 0.1471x_8 + 0.0169x_9 + 0.628x_{10} - 0.00869x_{11}$$

$x_1$ = driving distance in miles (distance)

$x_2$ = miles per kilowatt-hour (mpkWh)

$x_3$ = miles per gallon (mpg)

$x_4$ = electricity price per kWh (utildol)

$x_5$ = gasoline price per gallon (gasdol)

$x_6$ = distance x mpg

$x_7$ = distance x mpkWh

$x_8$ = distance x utildol

$x_9$ = distance x gasdol

$x_{10}$ = mpkWh x utildol

$x_{11}$ = mpg x gasdol

Fig.7 shows the result and the RMSPE is 0.1405. We also provide the results of the ordinary least-squares multiple regression model:

$$\hat{y} = -1.377 + 0.582x_1 - 0.197x_2 + 0.0523x_3 - 4.929x_4 + 0.591x_5 - 0.00222x_6$$
$$+ 0.00834x_7 - 0.204x_8 + 0.0241x_9 + 0.985x_{10} - 0.0131x_{11}$$

This model includes the main and interaction effect of the DOE parameters and the empirical driving distance, which accounts for 98% of the variance in the average difference. The RMSE is 0.262. Our approach identifies the same factors with a significant effect with the regression model, and our linear model is closer to the true function. This approach can be employed to address a discrete case, which expands the range of application.

**CONTROL STRATEGIES**

A control strategy is designed by utilizing this approximated GP. Considering a control scheme and given a reference output, our task is to repeatedly adjust the input values to enable the function to achieve the target. It is a two-step approach. With a training set, we learn the true underlying function with the approximated linear model from these training points. The training set can be obtained by observations from the control system. With the learned linear model, we determine the appropriate input values by solving a polynomial optimization program at each step. As previously introduced, the optimization program is

$$J = w_1(y^* - \hat{y})^2 + w_2(x - x')^2$$

$y^*$ = the reference value

$\hat{y}$ = the predictive value based on the current input

$x$ = the optimal input

$x'$ = the current input

At each step, we compute a predicted objective value $\hat{y}$ by the current input $x'$ to leverage the linear predictive model. Then, a polynomial optimization problem is solved and the best next-step input is the optimal solution. The optimization problems are sequentially solved until the target objective value is achieved. Next, we present several numerical examples of this control strategy.

a. Assume that the true function is

$$f(x) = x \sin x$$

$$x \in (-1,1)$$

After training, we learn the corresponding linear model:

$$\hat{y} = 0.4718\phi_0(x) + 0.3722\phi_2(x) - 0.0147\phi_4(x)$$

The reference value is set to 0.5, and the optimization cost function is

$$J = (y^* - \hat{y})^2 + (x - x')^2$$

The starting input point is $x_0 = 0.6$. As shown in Fig. 6, the process rapidly converges to the reference value. Because our linear model is an approximated model and a small gap is observed between the true objective value and our model's value, the reference value is not achieved; however, the optimum for the optimization polynomial is obtained. We employ a real-time adjustment trick to slightly change the linear model to enable the process to achieve the reference value.

[insert Fig. 7 here]

When the starting point is set to $x_0 = 0.8$ and the reference value is 0.1, the system operates as shown in Fig. 8.

[insert Fig. 8 here]

It shows that the approach is also suitable.

b. Assume that the true function is

$$f(x) = e^x$$

$$x \in (-1,1)$$

and the approximated linear model is

$$\hat{y} = 1.6305\phi_0(x) + 0.8690\phi_1(x) + 0.1942\phi_2(x)$$

We set the starting point to $x_0 = -0.5$ and the reference value to 1.5 and obtain the simulated control process in Fig. 9.

[insert Fig. 9 here]

c. We consider a bivariate case. We assume that the true function is the bivariate Gaussian distribution function

$$f(x) = \frac{1}{\sqrt{2\pi|\Sigma|}} \exp\left(-\frac{1}{2}(x-\mu)^T \Sigma^{-1}(x-\mu)\right)$$

and the parameters are

$$\mu = \begin{bmatrix} 0 \\ 0 \end{bmatrix}$$

$$\Sigma = \begin{bmatrix} 0.25 & 0.3 \\ 0.3 & 1 \end{bmatrix}$$

Because two variables exist, the optimization cost objective function is changed to

$$J = (y^* - \hat{y})^2 + \|x - x'\|_2^2$$

The initial point is $[-0.5 \quad 0.5]^T$, and the reference value is 1.0. The control process is shown in Fig. 10.

[insert Fig. 10 here]

Although the simulation is expensive and time consuming, the output value converges to the reference value in a few steps.

d. Then, we consider an example that is not suitable for this control strategy. Assume that the true function is

$$f(x) = x^3 - \frac{1}{2}x$$

$$x \in (-1,1)$$

and that the corresponding approximated linear model is

$$\hat{y} = 0.0868\phi_1(x) + 0.2107\phi_3(x)$$

Fig. 11 shows the plot of this function.

[insert Fig. 11 here]

This function is not monotonic. In the control scheme, the starting point is -0.7 and the reference objective value is 0.4.

[insert Fig. 12 here]

In the simulation process, as shown in Fig. 12, the function cannot attain the reference value. When the objective values are near the local maximum, the solutions to the optimization problems will converge to the local maxima. Then, we explore the convergence condition of our control strategy.

Let $x_0$ denote the starting point and let $x_k$ denote the optimal solution to the optimization problem

$$\min w_1(y^* - y)^2 + w_2(x - x_{k-1})^2$$

where $y = f(x)$ is the approximated linear model. Then, we obtain

$$w_1(y^* - y_k)^2 + w_2(x_k - x_{k-1})^2 \leq w_1(y^* - y_{k-1})^2$$

By the optimality condition,

$$-w_1(y^* - y_k)y_k' + w_2(x_k - x_{k-1}) = 0$$

Then,

$$w_1(y^* - y_k)^2 + \frac{w_1^2}{w_2} y_k'^{\,2} (y^* - y_k)^2 \leq w_1(y^* - y_{k-1})^2$$

$$\left(1 + \frac{w_1}{w_2} y_k'^{\,2}\right)(y^* - y_k)^2 \leq (y^* - y_{k-1})^2$$

$$\left(\prod_{j=1}^{n}\left(1 + \frac{w_1}{w_2} y_j'^{\,2}\right)\right)(y^* - y_k)^2 \leq (y^* - y_0)^2$$

The objective value converges to $y^*$ when

$$\lim_{n \to \infty}\left(\prod_{j=1}^{n}\left(1 + \frac{w_1}{w_2} y_j'^{\,2}\right)\right) = \infty$$

We can obtain a sufficient condition of the convergence in which the control process will converge to the reference objective value when the function between the initial point and reference point is strictly monotonic and $w_1 > 0$.

When the function is not strictly monotonic, we can change the control strategy slightly to enable us to compute an approximated optimal input by assigning a large value to $w_1$. When we assign the objective value around the reference value, we can adjust the input in the typical manner.

We perform the experiment with example d.

$$f(x) = x^3 - \frac{1}{2}x$$

$$x \in (-1,1)$$

We employ the optimization problem

$$J = 50(y^* - \hat{y})^2 + (x - x')^2$$

When the process is near the reference value, we change the optimization problem to

$$J = (y^* - \hat{y})^2 + (x - x')^2$$

The simulation results are shown in Fig. 13.

[insert Fig. 13 here]

Compared with the previous control process, the use of the modified strategy enables us to attain the reference value.

**CONCLUSIONS**

The GP is a powerful tool for learning functions from data due to its flexibility and consistency with computational tractability. However, it can be time consuming when it is applied to high-dimensional data. Thus, methods for approximating the GP, such as reduced-rank approximation of the Gram matrix (Fowlkes *et al.,* 2001) and greedy approximation, can be employed. These approximation methods can efficiently reduce the computation time. The disadvantage of these methods is that they cannot be integrated into a control strategy based on a GP. In a control strategy, we aim to calculate the best input at each time step to achieve the reference output value; the input value is difficult to compute with these approximation methods. Thus, we propose an approximation GP approach that can overcome these difficulties. It can efficiently conduct a GP regression and apply it to a control strategy.

In our approach, we leverage Legendre polynomials to generate a linear approximated model. This linear regression model can approximate a GP by utilizing a nonnegative garrote technique. Then, this linear approximated model can be employed as an objective function in a series of optimization problems to compute the best inputs to achieve the reference value in a control process. Several modeling and control examples are presented to demonstrate the efficacy of our approach. A sufficient condition is presented to ensure the convergence of this control strategy.

Our approximated regression model requires a trade-off of accuracy and efficiency. If we apply additional basis functions, a more accurate regression can be achieved but the efficiency will decrease as the number of parameters is increased. Otherwise, it will be efficient but the regression result may not be sufficiently accurate. When we use this approach, we should select the number of parameters as needed.

In the control strategy, we propose a two-step control method for cases in which the sufficiency condition is not satisfied. The use of a general method is not feasible because the scenario varies on a case-by-case basis. The general idea is that we should assign the process approach as the reference output value by the first-step control process and then obtain the exact point by the second-step control process.

**Figures**

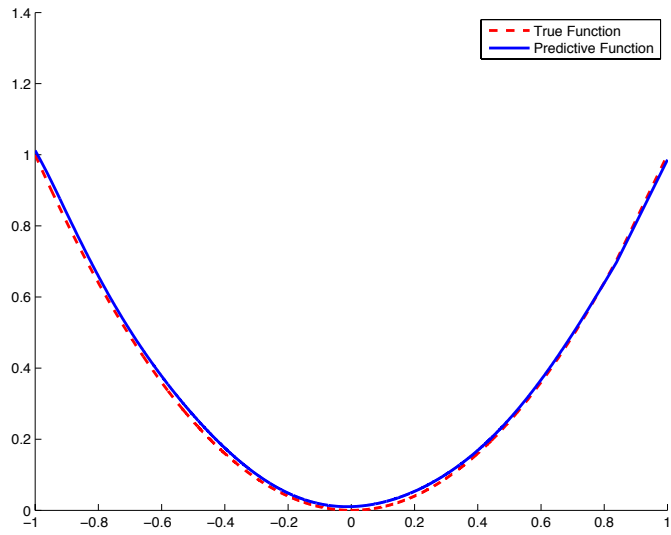

**Fig. 1.** True function and predictive function in univariate example a.

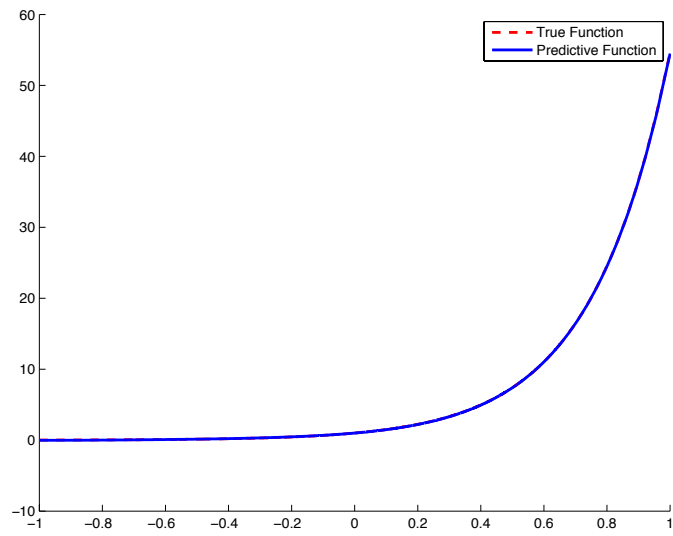

**Fig. 2.** True function and predictive function in univariate example b.

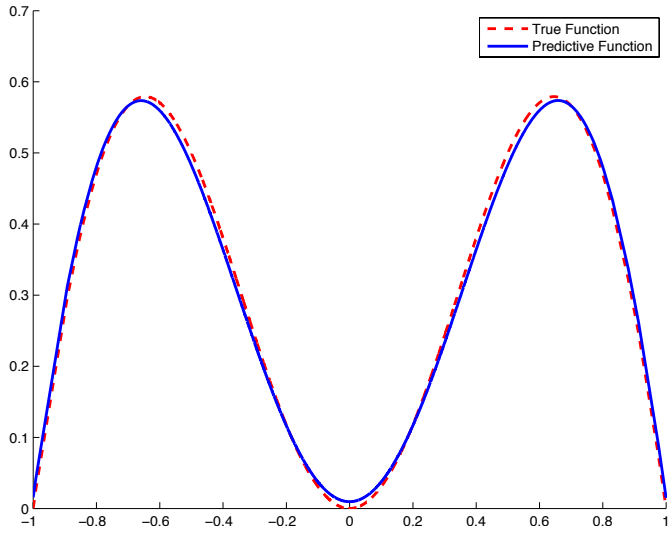

**Fig. 3**. True function and predictive function in univariate example c.

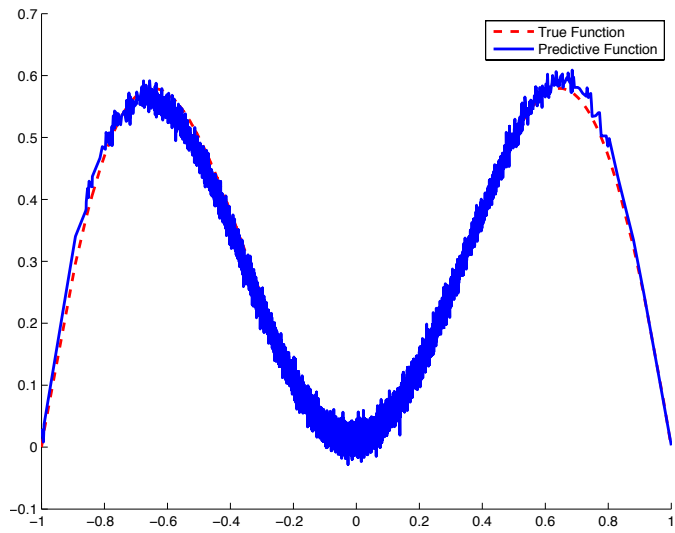

**Fig. 4.** True function and predictive function of original GP model in univariate example c.

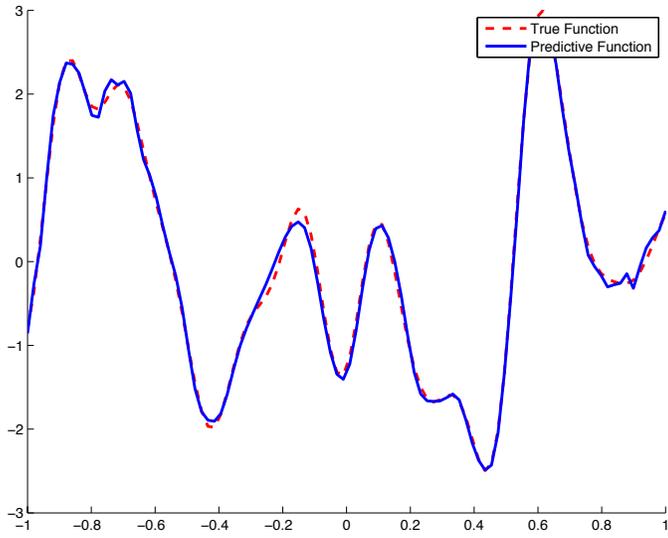

**Fig. 5.** True function and predictive function in univariate example d.

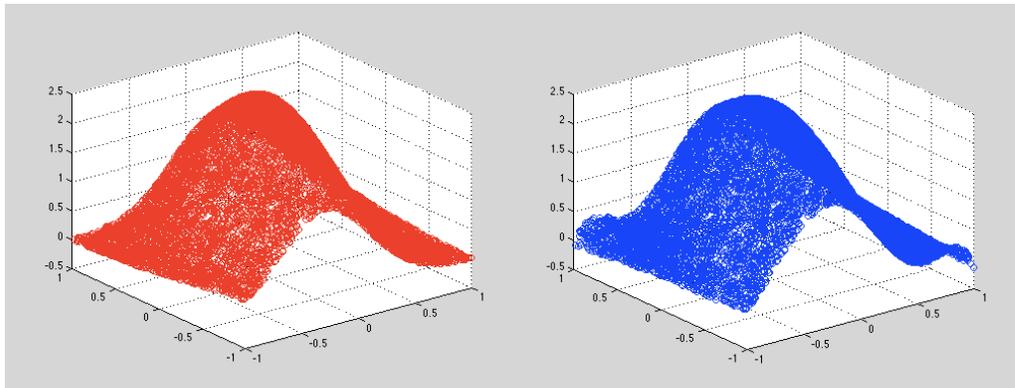

**Fig. 6**. True function and predictive function in the bivariate example.

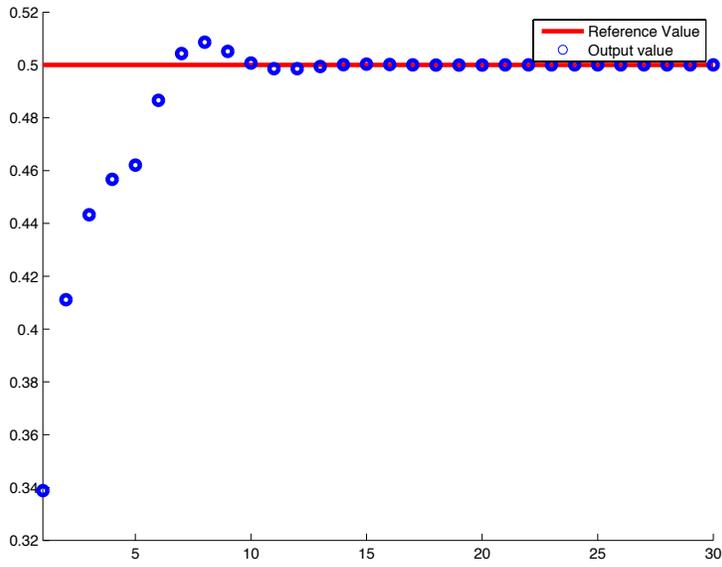

**Fig. 7.** First control process in univariate control example a.

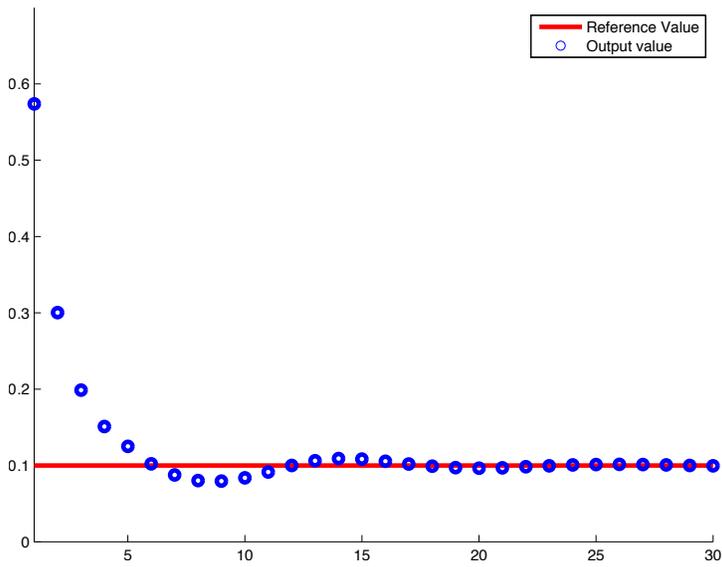

**Fig. 8.** Second control process in univariate control example a.

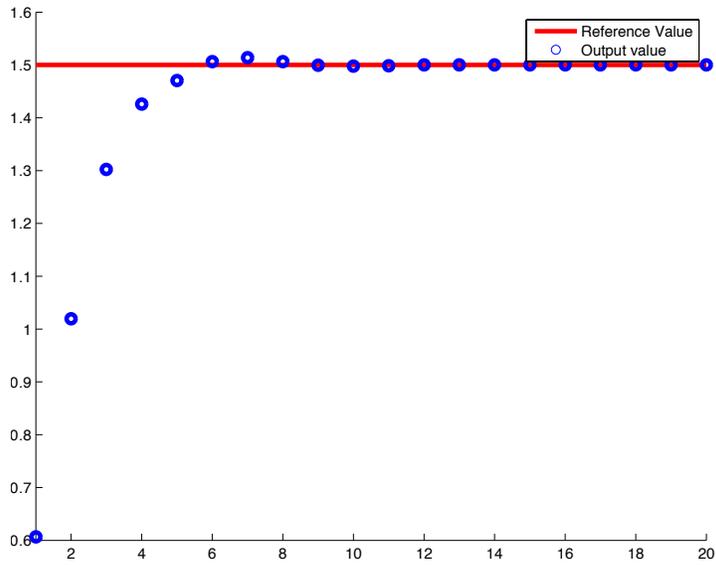

**Fig. 9.** Control process in univariate control example b.

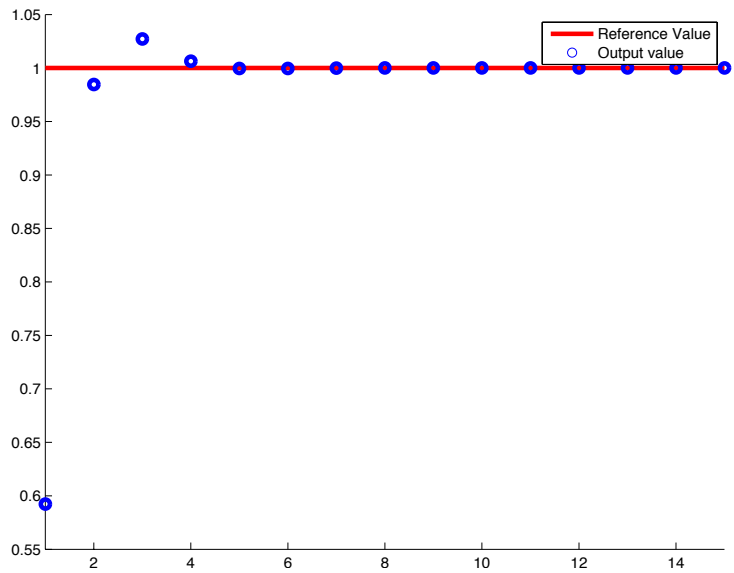

**Fig. 10.** Control process in the bivariate control example.

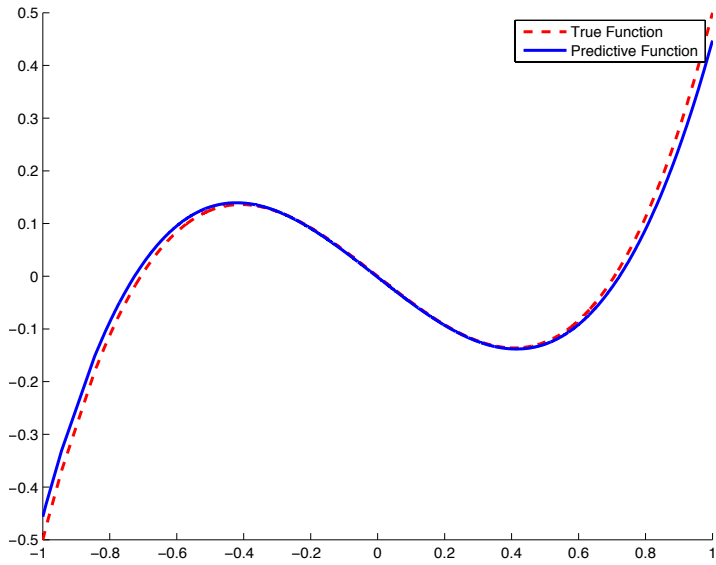

**Fig. 11.** True function and predictive function in control example d.

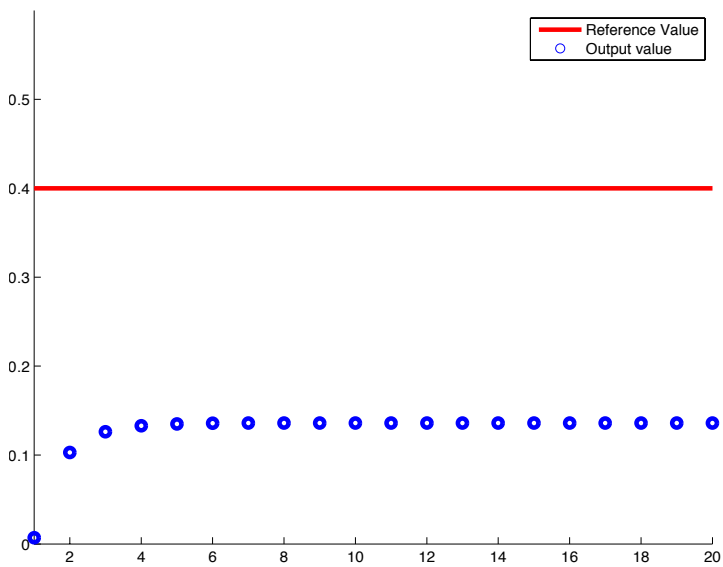

**Fig. 12.** Control process in control example d.

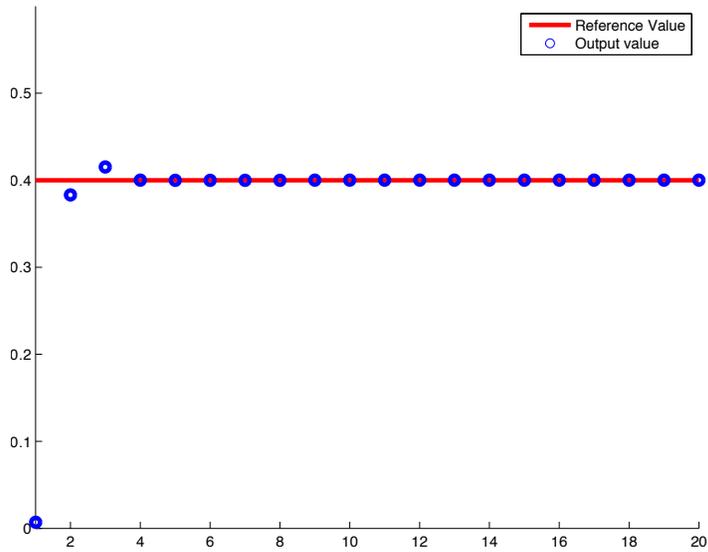

Fig. 13. Improved control process in control example d.